# A NEW SCHEME OF SIGNATURE EXTARCTION FOR IRIS AUTHENTICATION

[1] Belhassen AKROUT, [2] Imen KHANFIR KALLEL, [1] Chokri BENAMAR and [3] Boulbaba BEN AMOR

[1] REGIM: Research Group on Intelligent Machines Engineering National school of Sfax (ENIS), BP 1173, 3038 Sfax, Tunisia.
[2] ICOS: Intelligent Control, design Optimization of complex System, Sfax, Tunisia
[3] TELECOMLille1: Computer Science and Network Department, Lille, French
E-mail:akrout_belhassen@yahoo.fr, Chokri.BenAmar@enis.rnu.tn, boulbaba.benamor@telecom-lille1.eu, imen.khanfir@batotex.com.tn

## Abstract

*Iris recognition, a relatively new biometric technology, has great advantages, such as variability, stability and security, thus is the most promising for high security environment. Iris recognition is proposed in this report.*

*We describe some methods, the first one is based on grey level histogram to extract the pupil, the second is based on elliptic and parabolic HOUGH transformation to determinate the edge of iris, upper and lower eyelids, the third we used 2D Gabor Wavelets to encode the iris and finally we used the Hamming distance for authentication.*

## 1 Introduction

The security and the control of access always benefitted from a particular interest. This interest turned into obsession for some countries, notably following the evolution of the terrorist events. Immediately, several automatic systems of verification of identities have been put to days. A verification of identity can be either by identification is by authentication.

In the present work one is interested especially in the based systems of authentication on the recognition of the iris. This choice is based on the fact that this organ of the human being offers a big precision of discrimination. This approach of measure biometric was revealed one among the most reliable [10].

In this survey one proposes a strategy of recognition of the individuals by analysis of the picture of the iris. Such a strategy starts with the localization of the iris to succeed to the decision while passing by the extraction of the features and the classification of the data.

In this optics this paper articulates like continuation:

In first place one starts with exposing some reliable known techniques in the literature and in second place one describes the methods adopted for the detection of the internal and external contours of the iris followed of a description of the normalization.

The third stage is dedicated to the extraction of the biometric signature of the iris. This stage often had resort to an analysis multi-scale permitting the extraction of information contained in the zone of interest in the picture.

Fourth one describes the principle of coding and comparison.

One encloses one presents a general conclusion on the achieved system.

## 2 Related works

For localization of the iris DAUGMAN uses a strategy to determine the coordinates of the centers and the rays of the iris and the pupil. He applies operators of integration as detectors of contours, along a predetermined trajectory.

He assured that the contours of the pupil and the iris can be as circles but they cannot be concentric [3] [5]. He used an integro-differential operator to estimate three parameters for every circle: the coordinates of the center (ic, jc) and the r ray.

DAUGMAN [2] used multiscale quadrature wavelets to extract texture phase structure information of the iris to generate a 2048-bit iris code and compared the difference between a pair of iris representations by computing their Hamming distance to arrive to the authentication.

WILDES [13] uses a method to two stages to localize the iris: the detection of the side followed by transformed it of HOUGH. He converted, in the first place, the information of the intensity of the picture in a binary contour card and uses the directional detector then to discover the points of the side. This operation [1] [12] consists in a doorstep of the intensity of the picture gradient.

Then WILDES represented the texture of the iris with The Laplacian pyramid constructs with four levels of different resolutions and use the interrelationship normalized to determine if the picture of the entry and the picture models are the same class.

It is necessary to note that all these algorithms are based on pictures in levels of gray and all colorful information is not used.

## 3 Iris localization

The detection of the iris is a primordial stage in the system of authentication. It consists in drawing the internal contour, separate the iris of the pupil, as well as the external contour separates the iris of the sclera (white of eye) and the two lids.

### 3.1 Extraction of the pupil

Following the observation of the picture of eye one concluded that the pupil represents the darkest part. It is distinguished in the histogram of the gray levels by a peak of maximal intensity (Figure 1).

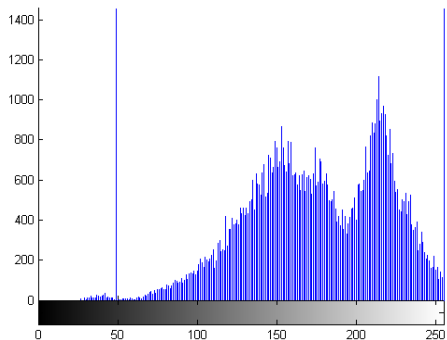

Figure 1: picture of the histogram that presents the peak of the pupil

One uses, to extract the pupil of its environment, a technique of histogram doorstep [8]. The doorstep is determined like continuation:
From the histogram, one looks for the indication of the gray level that possesses the biggest number of pixels.

This value is used then like a doorstep, to extract in a simple and fast manner the pupil of eye as the shows the figure 2.

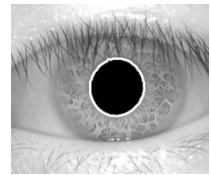

Figure 2: Result of detection of the contour of the pupil

The determination of the center of the pupil has a big importance in the continuation of the treatments. It is used like a landmark for all the shapes to detect (elliptic or parabolic) thus it is the center of the two circles that limits the zone susceptible to contain the contour of the iris.

The coordinates (x0, y0) of the center of the pupil are gotten following the application of the method of the barycenter that consists respectively to a projection of the pixels of the picture of the pupil on the axes of the abscissas. Such a projection permits to limit two segments.

The intersection of the two mediators of these two segments determines the coordinates of the center searched for of the pupil.

### 3.2 Determination of the external contour of the iris

The localization of the iris in space of picture requires the detection of external contour that can be gotten by the superposition of the elliptic shape and the parabolic shape.

These two shapes are gotten following the implementation of a method, based on the parametric models, which use some information a priori on the structure of the contour. It is about a simple and efficient method that carries the name of his inventor HOUGH [7].

#### 3.2.1 The research of the iris

The elliptic shape of the iris imposes to apply the elliptic transformation of HOUGH on a zone limited respectively by two concentric circles in the center of the pupil of coordinates (x0, y0) and ray 1,2 * r and 2,4 * r.

Such a transformation is composed therefore by primordial stages one is obliged to follow it. The first stage consists in preparing, from the picture in gray, a picture contours while applying the gradient followed of a threshold to make appear that the important information (the contours).

The second, it is a stage of sweep and transformation. She sums up as follows:
One makes browses it of the picture contour pixel by pixel while verifying if he contains a value of already fixed gradient superior or equal to the doorstep. If it is the case, it is necessary to get ready to a transformation from the space picture to the space accumulator.
Knowing that the equation of the ellipse writes itself as follows:

$$\begin{cases} x = a \cos(\theta) \\ y = b \sin(\theta) \end{cases} \quad (1)$$

The polar presentation of the ellipse is schematized like watch the figure 3.

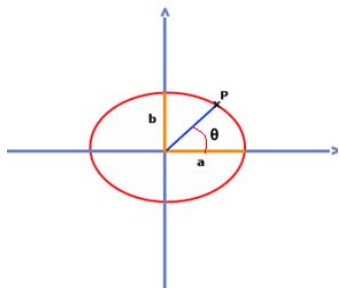

Figure 3: Presentation of an ellipse in polar systems coordinate

One can construct a space of coordinates (a, b, θ) while rewriting these equations according to x and y.

$$\begin{cases} a = (x - x_0) / \cos(\theta) \\ b = (y - y_0) / \sin(\theta) \end{cases} \quad (2)$$

Then for every pixel of coordinates (x, y), in the space of picture, one can find the elliptic representation there in the space accumulator (space (a, b, θ)) as fixing a and b and while varying θ of 0 to $2\pi$.

Our accumulator, of dimension 3D, is initialized to zeros. Every fixing of information (a, b, θ) is assured by an increment of the value of the places accumulator correspondent. The completion of this stage, produces an accumulator full of information on the ellipses to search for, that requires it resorts to a third stage to extract it.

One applies in the third stage a doorstep on our accumulator in order to minimize the time of treatment. For all slot that has a value passing the doorstep one looks for x and y, the coordinates of one point that possess a strong probability to belong to the border of the ellipse to search for, as fixing a and b and while varying θ, the figure 4 shows the result of detection.

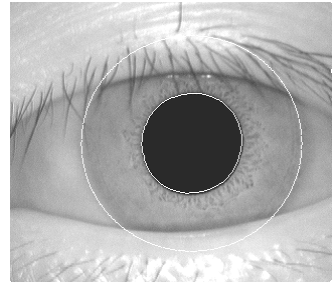

Figure 4 : Result of detection of the contour the iris-sclera

### 3.2.2 Detection of the contours iris - eyelid

For a human being, the iris is generally hidden partially by the two eyelids. It is useful to exclude them of the interest zone to extract the useful information of the iris that defers a man of the other only.

Indeed, our idea sums up in the detection of the contours of the two eyelids that is characterized by a parabolic and non elliptic shape [11].

The equation of a parabola writes itself in system of coordinates polar as follows:

$$\begin{cases} x = (d \cos(\theta)) / (1 + \cos(\theta)) \\ y = (d \sin(\theta)) / (1 + \cos(\theta)) \end{cases} \quad (3)$$

Where d is the distance between one point that belongs to the parabola and a point that belong to the normal and θ angle of variation.

In the goal to draw the two parabolas aims, one thought to search for to the minimum three points for every parabola a corresponding point to the summit and two other symmetrical points in relation to the axis of the ordinates.
Once again one proposes to use the method of HOUGH to find a set of points localized on the circumference of the parabolas searched for.

In first stage, one is interested in to determine a zone of research that varies according to the opening of the pupil and to determine the two parameters following every parabola:
d = distance between the center (x0, y0) and the lids.
θ = the angle of variation.

The used accumulator is bidimensional, (d, θ) initialized to zeros. One makes the course pixel previously by pixel of the definite research zone. For every pixel that passes a doorstep well one defines varies x and there to calculate the d and θ and there to assign some ways.

The following stage consists in applying a doorstep on the containing resulting accumulator the d and. to determine the points of the two parabolas belonging to the superior and lower contour. The set of these points is going to generate the two sought-after parabolas.

Finally, and like a result of our detection of the contours external of the iris, the pupil, the superior eyelid and the lower eyelid, the iris is localized well as it shows the figure 5.

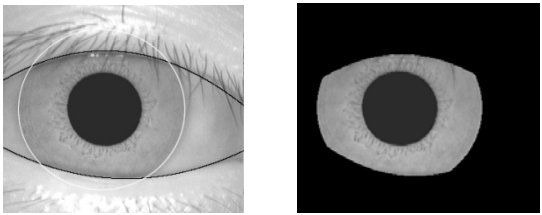

Figure 5: Result of detection of the contours external of the iris

## 4 Normalization of the interest zone

The size of the iris differs according to the plan of shot of the picture of eye or for the effect of retraction or dilation of the pupil. Indeed the variation and the change of the light intensity modify the size of the pupil (a strong intensity→ opening of the pupil, a weak intensity→ closing of the pupil). One finished whereas the size of the iris is not constant.

This distortion can influence the result of the extraction of the signature. One aims then to transform the picture of the iris in an oblong shape of stationary size (64 x 256 pixels), from where one is going to apply a pseudo transformation - polar [6].

$$I(x(r, \theta), y(r, \theta)) \rightarrow I(r, \theta) \quad (4)$$

With x (r, θ) and y (r, θ) are defined as the linear combinations of the points situated between the point (xp (θ), yp (θ) situated on the contour of the pupil and the point (xi (θ), yi (θ)) of the contour of the iris defined like follows :

$$x(r, \theta) = (1-r) * x_p(\theta) + r * x_i(\theta)$$
$$y(r, \theta) = (1-r) * y_p(\theta) + r * y_i(\theta) \quad (5)$$

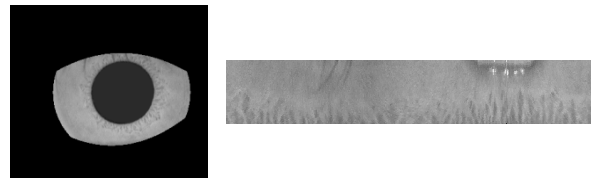

Figure 6: Normalization of the iris after detection

The resulting picture (figure 6) to a weak contrast and a luminance unbalanced because of the ambient light, what also acts on the result of the extraction and the comparison of the codes.

To correct this situation, one tried to heighten the histogram of the pseudo iris - polar while using a function of equating of the histogram. This function has for goal to modify the levels of gray of the picture to increase the contrast here under like watch the figure.

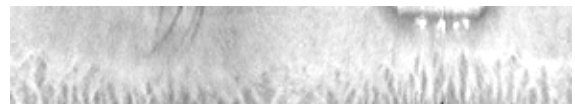

Figure 7: Result of the correction of the luminance

## 5 Extraction of the signature

The method of extraction of the signature is based on the implementation of transformed in bidimensional wavelet of Gabor.

Most waves use their capacity to approximate some classes of functions with a small number of non hopeless coefficients.

It is necessary to construct therefore Ψ in order to produce a maximum of coefficients of wavelet that is near of zero [9].

The bidimensional wavelet of Gabor obeys all these principles, and it has been used at the time of former works on the iris. This method is used in all systems present biometric based on the iris and it is proposed, developed and tested by DAUGMAN in 1997 to the University of Cambridge, England.

Transformed it in bidimensional wave of Gabor has the following shape:

$$\int_\rho \int_\phi e^{-i\omega(\theta_0-\phi)} e^{-(r_0-\rho)^2/\alpha^2} e^{-(\theta_0-\phi)^2/\beta^2} I(\rho,\theta)\rho \ d\rho \ d\phi \quad (6)$$

Where $I(\rho,\theta)$ represents the intensity of the picture in a polar coordinate. ω is the throbbing of the wavelet of Gabor, $r_0$ and $\theta_0$ are the coordinates of the point around of which one applies the wave, and finally α and β are the parameters of the analysis multi scale following the rays and the angles respectively.

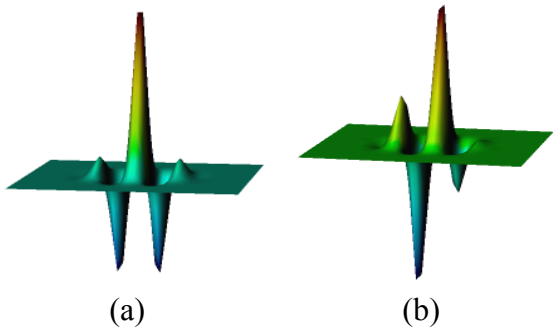

(a)          (b)

Figure 8: Real (a) and imaginary (b) shape of the wavelet of Gabor

We will use a basis of wavelet formed by a wavelet mother and three girls thus, inversely proportional to the parameters α and β that varies between 1.5 and 12. One analyzes the features of the iris following 64 points ($r_0$, $\theta_0$) that represent the centers of the wavelet.

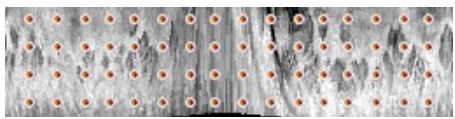

Figure 9: The disposition of the 64 centers of analyses

## 6 Coding of information

The application of every wavelet on a part of the iris will generate 2 bits according to the principle of the coding 4 quadrants. Let's notice that transformed it in wavelet possesses a complex shape, the principle of the coding 4 quadrants are illustrated in the face 10, and it consists in coding information contained in the phase of transformed it in wavelet according to his position in the trigonometric circle.

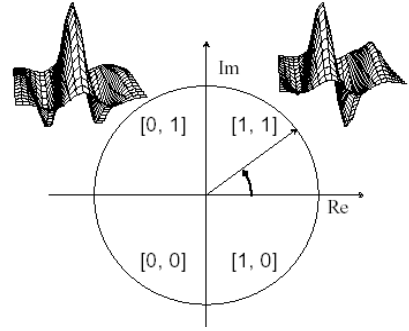

Figure 10: Principle of the coding 4 quadrants

Either Φ the phase of transformed it of Gabor 2D, and h codes it generated by this last applied to the picture.

If $0 < \Phi < \pi/2$    then h = [1,1]
If $\pi/2 < \Phi < \pi$    then h = [0,1]
                                                                  (7)
If $\pi < \Phi < 3\pi/2$    then   h = [0,0]
If $3\pi/2 < \Phi < 2\pi$    then   h = [1,0]

While applying this system of coding one gets a code of 2048 bits according to the method of the bidimensional wavelet of Gabor.

Information contained in the amplitude of transformed it in wavelet of the picture is not very meaningful and very dependent of the outside factors as illumination, the contrast and the gain of the camera [4].

## 7 Comparisons

One explained previously how to extract a code of 256 bytes from the iris. Now, it is necessary to think about a technique that facilitates us the comparison of two codes in order to identify the impostors of authenticated them.

The comparison bit to bit of every two pair of codes of iris A and B is given by the normalization of the distance of HAMMING (HD) [5] definite below as follows:

$$HD = 1/N \sum_{j=1}^{N} C_A(j) \oplus C_B(j) \quad (8)$$

With $\oplus$ is the operator Boolean (XOR) describes as follows:

$$S = a \otimes b = \overline{a}b + a\overline{b} \quad (9)$$

The famous aray 1 the results possible of comparison of two bits with the operator with have a bit of the code A and b is the corresponding bit of the code B.

| a | b | $\bar{a}$ | $\bar{b}$ | $a\bar{b} + \bar{a}b$ | S |
|---|---|---|---|---|---|
| 0 | 0 | 1 | 1 | 0 | 0 |
| 0 | 1 | 1 | 0 | 1 | 1 |
| 1 | 0 | 0 | 1 | 1 | 1 |
| 1 | 1 | 0 | 0 | 0 | 0 |

Array 1: Results possible of application of XOR on two bits

The result of application of the operator Boolean (XOR) on 2 bits equals to 1 if and only if the two bits A (j) and B(j) are different.

The probability so that 2 bits of any code of iris are equal or no is worth P = 0, 5.

When the majority of the costs A (j) and B (j) are equal, one is therefore in the case or the two codes compared are equivalent. In the inverse case, the two codes are different.

Generally, when HD stretches toward zero the species is authenticated and in the case where HD stretches toward one, the species is an impostor.

# 8 problems of comparison and the new solution proposed

## 8.1 problems of comparison

The angle of capture of eye depends on the position and the degree of rotation of the head what poses for us problems of authentication even though the compared irises belong to the same eye.

One presents in the continuation a solution that tempts to get round this problem that is based on the technique of correction of the angle of rotation of eye.

## 8.2 The new solution proposed

In the stages of detection of the contours of the two superior and lower lids, and after observation of eye, one concluded that the extreme tips left and right of eye is invariable (figure 11).

One can use this observation then to develop a technique using the coordinates $(x_{p1}, y_{p1})$ that represent respectively the intersection of the two parabolas of the superior and lower lids and to adjust the positioning of l then eye.

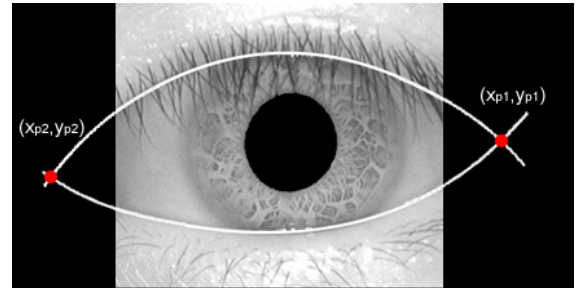

Face 11: The two tips of eye

The coordinates of the two points P1 and P2 are gotten following the determination of the equations respective of the superior and lower parabola. One:

$a_1X^2 + b_1X + c_1$
(Equation of the first parabola)

$a_2X^2 + b_2X + c_2$
(Equation of the second parabola)
(10)

$A_1X^2 + b_1 X + c_1 = a_2 X^2 + b_2 X + c_2$

$=> X^2*(a_1 - a_2) + X*(b_1 - b_2) + (c_1 - c_2) = 0$

The resolution of this equation gives us $x_{p1}$ and $x_{p2}$ and while replacing $x_{p1}$ in the first equation and $x_{p1}$ in the second, one finds $y_{p1}$ and $y_{p2}$ respectively.

One now wants to make revolve l' .il around the center $(x_0, y_0)$ of the iris (figure 12). For it, one takes one of the points $(x_{p1}, y_{p1})$ or $(x_{p2}, y_{p2})$ and the center of coordinates $(x_0, y_0)$.

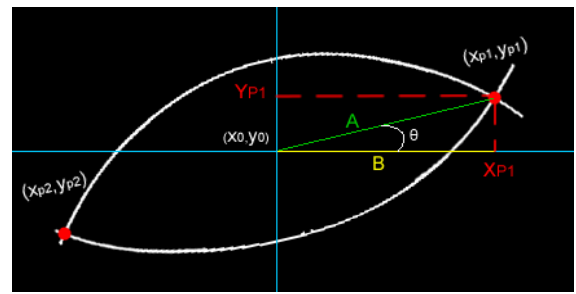

Figure 12: Determination of θ the angle of rotation

Taking A the distance between the center of the iris and the point $(x_{p1}, y_{p1})$, B outdistances it between the center and the projection of the point $(x_{p2}, y_{p2})$ on the axis of the abscissas and θ the angle of sought-after rotation.

$A = \sqrt{(x_{p1} - x_0)^2 + (y_{p1} - y_0)^2}$
$B = | x_{p1} - x_0 |$ (11)
$\cos(\theta) = A/B => \theta = \arccos(A/B)$

θ will be the angle of rotation of the picture, and for every value of θ corresponds a very determined position of eye.

## 9 histogram of the authentication

Once the distance of HAMMING is acquired us can get the doorstep of decision given by the intersection of the curves associated to the authentic and to the impostors (figure13).

We drew these two curves and we got a value of doorstep equal to 0, 39.

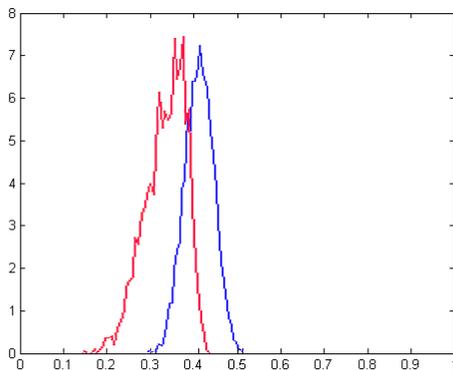

Figure 13: Doorstep of decision

## 10 conclusions

The present use of the authentication technology to basis of iris is limited because it is expensive.It is localized in some airports reappointed in the world to assure the security of the nations.
The perspectives of this authentication system by detection of iris depend on the applications but it can be spilled in highly secured other daily domains as the management of the accounts and the strong cases in the banks, the secured computers and in the luxurious cars.